\documentclass[10pt,twocolumn,letterpaper]{article}

\usepackage[pagenumbers]{cvpr} 

\usepackage{colortbl}
\usepackage[dvipsnames]{xcolor}
\usepackage{graphicx}
\usepackage{booktabs}
\usepackage{wrapfig}
\usepackage{multicol}
\usepackage{multirow}
\usepackage{pifont}

\newcommand{\titletext}{{OccFeat: Self-supervised \underline{Occ}upancy \underline{Feat}ure Prediction\\ for Pretraining BEV Segmentation Networks}}

\newcommand{\method}{{OccFeat}}

\newcommand{\effbzero}{{EN-B0}}
\newcommand{\resnetfifty}{{RN-50}}

\newcommand{\alsobaseline}{{Img-ALSO}}

\newcommand{\image}{I} 
\newcommand{\imagefeat}{F} 

\newcommand{\imageenc}{E_\text{I}}
\newcommand{\projector}{P_\text{B}}
\newcommand{\decoder}{D_\text{B}} 

\newcommand{\lifting}{D_\text{V}}

\newcommand{\teacherenc}{\imageenc^\text{Y}}

\newcommand{\bevfeaturemap}{F_\text{B}} 
\newcommand{\bevfeaturvolume}{F_\text{V}}

\newcommand{\teacherfeat}{Y}

\newcommand{\losstot}{\mathcal{L}}
\newcommand{\lossrec}{\mathcal{L}_\text{occ}}
\newcommand{\lossdistill}{\mathcal{L}_\text{feat}}

\newcommand{\voxgrid}{\mathcal{V}}

\newcommand{\occ}{O}
\newcommand{\occvox}{\mathcal{V}_{\text{Occ}}}

\newcommand{\numchannels}{N_\text{B}}
\newcommand{\bevheight}{H_\text{B}}
\newcommand{\bevwidth}{W_\text{B}}
\newcommand{\bevzed}{Z_\text{B}}
\newcommand{\numdistillchannels}{N_\text{y}}

\newcommand{\realnum}{\mathbb{R}}

\def\Plus{\texttt{+}}
\def\Minus{\texttt{-}}

\newcommand{\better}[1]{{\textcolor{ForestGreen}{(#1)}}}
\newcommand{\worse}[1]{{\textcolor{BrickRed}{(#1)}}}

\newcommand{\cmark}{\ding{51}}%
\newcommand{\xmark}{\ding{55}}%

\definecolor{baselinecolor}{rgb}{0.9, 0.9, 1.0}
\newcommand{\baseline}[1]{\cellcolor{baselinecolor}{#1}}

\newcommand{\parag}[1]{\smallskip\noindent\textbf{#1}\enspace}

 \usepackage{multirow}
 \usepackage[accsupp]{axessibility} 

\definecolor{cvprblue}{rgb}{0.21,0.49,0.74}
\usepackage[pagebackref,breaklinks,colorlinks,citecolor=cvprblue]{hyperref}

\def\paperID{14} 
\def\confName{CVPR}
\def\confYear{2024}

\title{\titletext}

\author{%
Sophia Sirko-Galouchenko\textsuperscript{1,2} 
\and
Alexandre Boulch\textsuperscript{1} 
\and
Spyros Gidaris\textsuperscript{1}
\and
Andrei Bursuc\textsuperscript{1}
\and
Antonin Vobecky\textsuperscript{1,3,4}
\and
Patrick P\'erez\textsuperscript{5}\thanks{Work done at valeo.ai.}
\and
Renaud Marlet\textsuperscript{1,6}
\vspace{1mm}
\and
\textsuperscript{1}Valeo.ai, Paris, France \hspace{1.2mm} 
\textsuperscript{2}ISIR, Sorbonne Universit\'e, Paris, France \hspace{1.2mm}
\textsuperscript{3}FEE CTU\thanks{Czech Institute of Informatics, Robotics and Cybernetics at the Czech Technical University in Prague.}, Prague, Czechia
\textbf{}
\vspace{1mm}
\\
\textsuperscript{4}CIIRC CTU, Prague, Czechia\hspace{1.2mm} 
\textsuperscript{5}Kyutai, Paris, France
\textbf{}
\vspace{1mm}
\\
\textsuperscript{6}LIGM, Ecole des Ponts, Univ Gustave Eiffel, CNRS, Marne-la-Vall\'ee, France
}

\begin{document}
\maketitle
\begin{abstract}
We introduce a self-supervised pretraining method, called \method, for camera-only Bird's-Eye-View (BEV) segmentation networks.
With \method, we pretrain a BEV network via occupancy prediction and feature distillation tasks. 
Occupancy prediction provides a 3D geometric understanding of the scene to the model. However, the geometry learned is class-agnostic. Hence, we add semantic information to the model in the 3D  space through distillation from a self-supervised pretrained image foundation model. 
Models pretrained with our method exhibit improved BEV semantic segmentation performance, particularly in low-data scenarios. 
Moreover, empirical results affirm the efficacy of integrating feature distillation with 3D occupancy prediction in our pretraining approach.

  \end{abstract}    
\section{Introduction}
\label{sec:intro}

Camera-only bird's-eye-view (BEV) networks have gained significant interest in recent years within the field of autonomous driving perception~\cite{lss, simplebev, huang2021bevdet, li2022bevformer, hu2021fiery, bartoccioni2023lara, zhou2022cross}. 
The appeal of the BEV, or else top-view, is that it 
offers a unified space for various sensors, including surround-view cameras, Lidar, and radar~\cite{liu2023bevfusion,hendy2020fishing,man2023bev}, for both annotation and runtime perception purposes, and can serve as input for subsequent tasks in the driving pipeline, such as forecasting and planning~\cite{hendy2020fishing,hu2021fiery,akan2022stretchbev,buhet2020plop, hu2023planning}.
Common tasks in the BEV space are semantic segmentation of objects~\cite{lss, simplebev, bartoccioni2023lara, zhou2022cross} and layouts~\cite{li2022hdmapnet, li2023bimapper},
as well as object detection~\cite{huang2021bevdet, li2022bevformer}. Our work specifically targets the camera-only BEV semantic segmentation task.

Until now, training networks for camera-only semantic segmentation in BEV space has relied on full supervision, necessitating annotations for each scene. This process is time-consuming due to the transition from the input image to the ``synthetic'' BEV space. For instance, annotations are typically generated for Lidar data, checked for visibility and classes in images, and then projected onto BEV segmentation images. To reduce annotation costs, we explore the potential of self-supervised pretraining camera-only BEV segmentation networks.

In self-supervised pretraining, networks are typically trained on annotation-free pretext tasks before the primary (downstream) task, like semantic segmentation.  
The aim is to guide the network in learning useful data representations during this pretraining phase. This process is intended to enhance the network's performance on the downstream task, enabling it to achieve higher accuracy while utilizing a reduced amount of annotated data.
Pretraining has been proven efficient for several modalities from images~\cite{chen2020simple, he2022masked} to Lidar~\cite{also} and with different strategies, such as contrastive learning~\cite{chen2020simple, he2020momentum}, teacher-student architectures~\cite{gidaris2021obow, gidaris2023moca, grill2020bootstrap, dino} or reconstruction tasks~\cite{he2022masked, bao2022beit}.
\begin{figure}
\makeatletter
\makeatother
\centering
\includegraphics[width=1.0\linewidth]{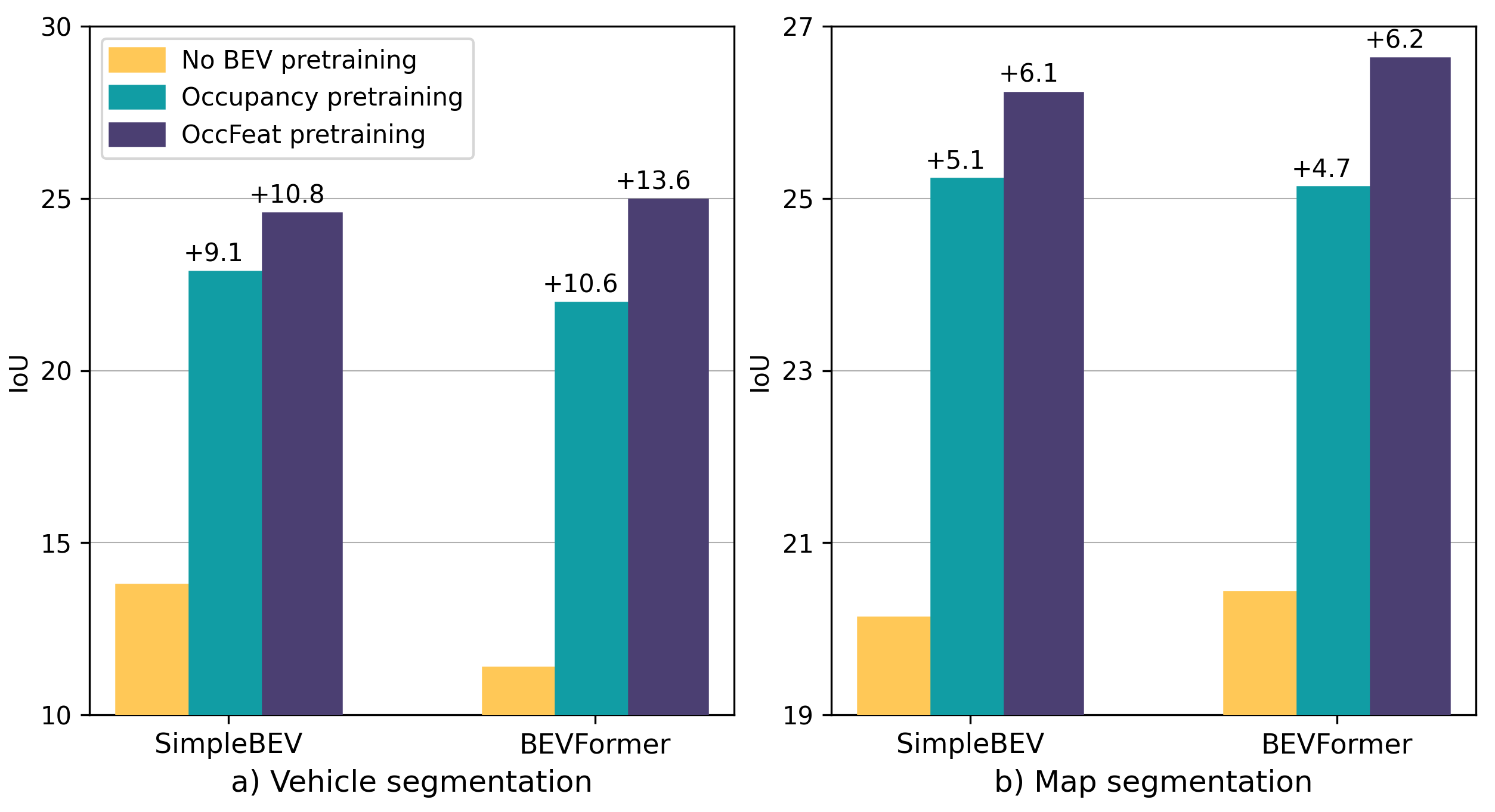}
\vspace{-22pt}
\caption{Performance comparison in low data regime (1\% annotated data of nuScenes)}
\label{fig:teaser}
\vspace{-10pt}
\end{figure}

In the field of autonomous driving, self-supervised pretraining for camera-only BEV networks has received limited attention despite its crucial role. 
Recently, a few methods have emerged that delve into this subject~\cite{yang2024visual, min2024multi, yang2024unipad}, but they predominantly focus on pretraining with 3D geometry prediction tasks.
For instance, ViDAR~\cite{yang2024visual} employs Lidar point cloud forecasting for pre-training, UniPAD~\cite{yang2024unipad} 3D surface and RGB pixels reconstruction, and UniScene~\cite{min2024multi} 3D occupancy prediction. While these methods equip the BEV networks with 3D geometry understanding, they often fall short in making the network capture semantic-aware information of the 3D scene, essential for tasks like BEV-based semantic segmentation.

Our approach, called Occupancy Feature Prediction (\method), addresses this gap by presenting a pretraining objective that promotes a more comprehensive understanding of the 3D scene, encompassing both geometric and semantic aspects.
In our approach, the camera-only BEV network is tasked to predict a 3D voxel-grid representation that includes (a) features indicating voxel occupancy and (b) high-level self-supervised image features characterizing occupied voxels.

To create this target voxel grid representation, we leverage aligned Lidar and image data in autonomous driving setups, along with a self-supervised image foundation model like DINOv2~\cite{dinov2}, which has been pretrained to extract high-level 2D image features. Specifically, the occupancy of each voxel is determined using Lidar data, considering a voxel occupied if it contains at least one Lidar point. Simultaneously, the self-supervised image foundation model fills the occupied voxels with high-level image features. This process involves projecting the center coordinates of each occupied voxel cell into the 2D space of the image features extracted from the foundation model.

Unlike approaches solely focused on 3D geometry prediction,
our method goes beyond by training the BEV network to predict a richer, more semantic representation of the 3D scene, all without requiring manual annotation, leveraging the pre-trained image foundation model. 
We empirically demonstrate that this enhancement leads to significantly better downstream BEV semantic segmentation results, especially in low-data regimes (e.g., see \cref{fig:teaser}).

Our contributions are the following:
\begin{enumerate}
    \item We present \method, a self-supervised pretraining approach for camera-only BEV segmentation networks that enforces both geometric and semantic understanding of 3D scenes.
   
    \item \method~exploits three modalities for pretraining: image, Lidar, and DINOv2 features. To the best of our knowledge, we are the first to leverage foundation image models (DINOv2) for pretraining camera-only BEV networks. 
    We note that after pretraining, the Lidar and DINOv2 data are not used anymore.

    \item We evaluate \method~on nuScenes~\cite{nuscenes} for BEV semantic segmentation of both vehicles and map layout. The results show the benefit of our pretraining method, especially in low-shot regimes, e.g., when using annotations only for 1\% or 10\% of nuScene's training data.
    Additionally, our \method~pretraining improves the robustness, as evaluated on the nuScenes-C benchmark~\cite{xie2023robobev}.

\end{enumerate}

\section{Related work}
\label{sec:related}

\parag{Camera-only BEV perception.} 
BEV perception aims for a unified representation of the surrounding environment of a vehicle. BEV has recently arised as a prevailing paradigm for multi-camera perception systems for autonomous driving. 
Camera-based BEV models are typically composed of three parts: (i) an image encoder shared across cameras for extracting 2D features, (ii) a view-projection module for ``lifting'' features in the 3D space 
to produce BEV features, and (iii) one or more task decoder modules that process BEV features towards addressing a task of interest, e.g., semantic segmentation, map prediction, 3D detection, etc. Among them, the view-projection has been in the spotlight of numerous works in this area in the past few years. The diversity of the approaches for view projection is impressive, ranging from purely geometric ones, e.g.,  using inverse perspective mapping with strong assumptions about the world~\cite{sengupta2012automatic} 
to projections completely learned from data, typically leveraging a cross-attention mechanism between image features (often imbued with 3D knowledge or priors) and learnable-queries from the projected space~\cite{bartoccioni2023lara, zhou2022cross,liu2022petr,liu2023petrv2,xiong2023cape}.
Currently, the most commonly used ones belong to one of the so-called ``push'' or ``pull'' BEV projection approaches~\cite{simplebev, li2023fb}. 
The former are derived from the seminal Lift-Splat-Shoot (LSS)~\cite{lss} that leverages depth uncertainty estimations from each view to project features in a shared BEV space. The predictive performance of LSS can be further improved with other supervision signals, e.g., depth from Lidar points~\cite{reading2021categorical,li2023bevdepth},
stereo from time~\cite{li2023bevstereo, li2023bevstereo++,wang2022sts} 
or different depth parameterization~\cite{yang2023parametric}. 
Runtime speed can be significantly reduced thanks to custom pooling strategies~\cite{huang2021bevdet,liu2023bevfusion, huang2022bevpoolv2,zhou2023matrixvt,li2023fast,xie2022m}. 
\textit{Pull} approaches forego depth estimation and, instead, map 3D locations from the BEV space to the image space of the cameras. Then, they collect image features with deformable attention~\cite{li2022bevformer} or bilinear interpolation~\cite{simplebev} and spread them in the 3D space filling it. Pull approaches have a  simpler projection and have also been largely adopted and improved~\cite{yang2023bevformer,jiang2023polarformer,chen2022polar}. 
While predictive performance of BEV methods has been improved significantly in the last two years, very few methods deal with annotation-efficient learning beyond data augmentation strategies in the image or BEV space~\cite{huang2021bevdet, li2023fast}. 
Obtaining precise 3D annotations for BEV perception is a costly multi-stage labour-intensive process involving annotation of both point clouds and images from multiple cameras.
In this work we propose a strategy to improve the per-annotation efficiency of different BEV models and showcase it on two types of view-projections: SimpleBEV~\cite{simplebev} and BEVFormer~\cite{li2022bevformer}.

\parag{Self-supervised representation learning }
is a prominent paradigm that leverages unlabelled data towards producing useful representations (generalizable, robust, ready-to-use)
for different tasks of interest. Self-supervised learning (SSL) defines an annotation-free \textit{pretext task} that is determined solely by raw data, with the aim of providing supervision signal to extract useful patterns from the data and to learn representations. SSL pretrained models are subsequently used \emph{off-the-shelf}, probed or finetuned on different tasks of interest with limited amounts of labels, displaying better performance per-annotation compared to fully supervised counterparts~\cite{henaff2020data}. In the image domain, a myriad of SSL pretext tasks have been proposed such as predicting perturbations incurred into an image~\cite{gidaris2018unsupervised, doersch2015unsupervised, zhang2016colorful},
contrastive learning by separating similar from dissimilar views~\cite{he2020momentum, chen2020simple,misra2020self, chen2021exploring}, 
learning by clustering~\cite{caron2018deep,caron2020unsupervised}, 
self-distillation~\cite{grill2020bootstrap, gidaris2021obow,dino,bardes2022vicreg, assran2023self},
or most recently masked-image modeling objectives, particularly suitable for Vision Transformer architectures~\cite{bao2022beit, gidaris2023moca, dinov2, he2022masked, zhou2022ibot}. 
In spite of the success on curated internet images, e.g., ImageNet, SSL pretraining of image backbones on driving data is non-trivial, due to the high redundancy and class imbalance specific to driving data~\cite{chen2021multisiam, xiong2021self}. 
However, the availability of multiple sensor types on the vehicle, such as Lidar, and the emergence of vision foundation models, opens the door to different strategies to acquire 3D and/or semantic knowledge into image encoders for driving perception, e.g., unsupervised semantic segmentation~\cite{vobecky2022drive} 
or detection~\cite{tian2021unsupervised}. 
The synchronization of surround-view cameras and Lidar is also leveraged for pretraining Lidar networks with label-free knowledge distillation from pretrained image models leading to substantial performance gains in low-label regimes~\cite{sautier2022image, mahmoud2023self, liu2023segment, puy24scalr} 
and generalization~\cite{puy24scalr}.

\parag{BEV pretraining and distillation.}
Pretraining is commonly used for perception in autonomous driving (2D/3D object detection, semantic segmentation) to boost performance and compensate for the scarcity of labeled data and task difficulty, requiring semantic and 3D geometry awareness. ImageNet or depth estimation pretraining are widely used~\cite{park2021pseudo} for monocular perception. For BEV perception most works start from a backbone pretrained for monocular 3D object detection~\cite{wang2021fcos3d}. The shared BEV space between cameras and Lidar enables new forms of pretraining by distilling 3D reasoning skills from Lidar networks into camera-based ones in order to compensate for potential loss of geometric information in the view-projection module~\cite{huang2023geometric, hong2022cross, huang2022tig, wang2023distillbev,chen2023bevdistill,liu2023geomim}. They take the form a teacher-student architecture, with the Lidar teacher network trained in a supervised manner on 3D annotations which are however costly to acquire. First approaches to pretrain camera-based BEV networks following the SSL paradigm with an annotation-free pretext task, such as occupancy estimation~\cite{min2024multi}, forecasting Lidar point clouds~\cite{yang2024visual}, reconstructing 3D surfaces and RGB images~\cite{yang2024unipad}, have emerged only very recently with promising results. However they focus mostly on purely geometric cues or rendering RGB pixels which can be sub-optimal for downstream perception performance~\cite{balestriero2024learning} and potentially hinder the existing semantic knowledge in the image encoder. We propose to distill pretrained image features from DINOv2~\cite{dinov2} in the voxel space such that the produced BEV features are not only geometry aware, but also semantic aware. Closest to our work, is the recent POP-3D~\cite{vobecky2023pop} that distills CLIP features~\cite{radford2021clip} into a 3D occupancy prediction model~\cite{huang2023tri} towards open-vocabulary perception and not for pretraining.

\section{Method}
\label{sec:method}

\begin{figure*}[t]
    \centering
    \includegraphics[width=0.90\linewidth]{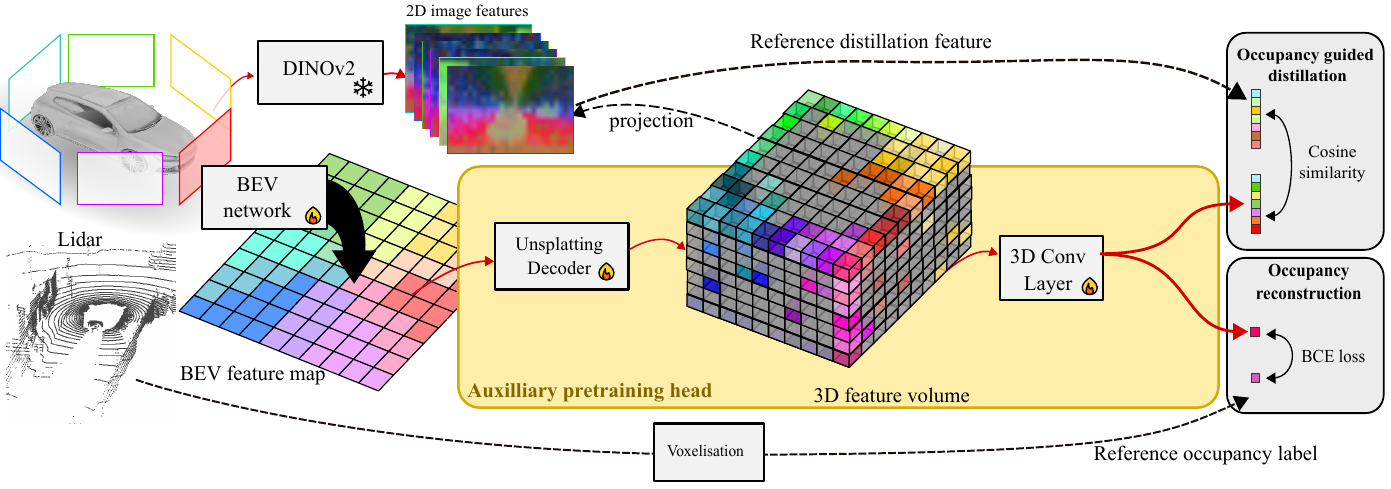}
    \vspace{-10pt}
    \caption{\textbf{Overview of \method's self-supervised BEV pretraining approach.}
    \method~attaches an auxiliary pretraining head on top of the BEV network. This head ``unsplats'' the BEV features to a 3D feature volume and predicts with it (a) the 3D occupancy of the scene (occupancy reconstruction loss) and (b) high-level self-supervised image features characterizing the occupied voxels (occupancy-guided distillation loss). 
    The occupancy targets are produced by ``voxelizing'' Lidar points (see \cref{fig:occ_voxels}), while the self-supervised image foundation model DINOv2 provides the feature targets for the occupied voxels. The pretraining head is removed after the pretraining. }
    \label{fig:pipeline}
    \vspace{-10pt}
\end{figure*}

Our goal is to pretrain a camera-only 
BEV segmentation network in a self-supervised way.
To this end, we intend
to equip the learned BEV representations with the ability to encode both the 3D geometry of the scene and semantic-aware information, crucial for downstream tasks such as semantic segmentation within the BEV space.
To achieve this goal, we leverage the availability of (i) aligned Lidar and image data in autonomous driving setups and (ii) a self-supervised pretrained image encoder able to extract high-level 2D features from images (e.g., DINOv2~\cite{dinov2}).
The proposed self-supervised BEV pretraining method \method, illustrated in Figure~\ref{fig:pipeline}, encompasses two training objectives:

\begin{description}
    \item[Occupancy reconstruction ($\lossrec$):] 
    This objective enforces the BEV network to capture the 3D geometry of the scene through an occupancy reconstruction task, defined using the available Lidar data.
   
    \item[Occupancy-guided feature distillation ($\lossdistill$):] 

    This objective enforces the BEV network to reconstruct high-level semantic features. The network is trained to predict, at occupied voxel locations, the features of an off-the-shelf self-supervised pretrained image encoder.
    
\end{description}
The total objective that our self-supervised pre-training approach minimizes is:
\begin{equation} \label{eq:total_loss}
\losstot = \lossrec + \lambda \cdot \lossdistill,
\end{equation}
where $\lambda$ is the weight coefficient for balancing the two loss terms. Except otherwise stated, we use $\lambda=0.01$.   

In the following, we begin with a brief overview of camera-only BEV networks in \cref{sec:bev_overview}. Then, we describe our occupancy reconstruction objective in \cref{sec:occupancy_reconstruction} and our occupancy-guided feature distillation objective in \cref{sec:feature_distillation}.

\subsection{BEV networks} \label{sec:bev_overview}

The BEV networks aim to build a BEV feature map from registered image data. In our case, this feature map is used for semantic segmentation.
These networks share a common architecture composed of 1) an image encoder, 2) a projection module and 3) a decoder.

\noindent{\textbf{The image encoder}}
$\imageenc$ produces a feature map $\imagefeat_c$ for each image $\image_c$, with $c$ ranging from $1$ to $C$, where $C$ is the total number of surround-view cameras in the vehicle.
Typically, these encoders come either from ResNet~\cite{resnet} or EfficientNet~\cite{effnet} family of models.

\noindent{\textbf{The projection module}}
$\projector$ is the module responsible for changing the representation space, from the sensor coordinate system (image features $\imagefeat_c$) to the BEV space.
$\projector$ takes as input the image features $\{\imagefeat_c\}_{c=1}^{C}$ and camera calibration and projects the image features in the BEV space.
Architectures differ in the way they operate this projection, from a full image feature volume aggregated over the vertical axis (SimpleBEV \cite{simplebev}) or a sparser volume filled according to an estimated depth distribution 
(LSS~\cite{lss}) to an attention-based projection as in CVT~\cite{zhou2022cross}, BEVFormer~\cite{ li2022bevformer}.

\noindent{\textbf{The decoder}}
$\decoder$ takes as input the image features in the BEV space generated by the $\projector$ and
further processes them with 2D convolutional layers and optionally upsamples them to the desired segmentation map resolution~\cite{lss,simplebev}.

This produces the output BEV features $\bevfeaturemap \in \realnum^{\numchannels \times \bevheight \times \bevwidth}$, where $\bevheight \times \bevwidth$ is the spatial resolution of the BEV features and $\numchannels$ is the number of feature channels.

\parag{Architecture-agnostic BEV representation pretraining.}
The self-supervised pretraining approach that we propose is applied on these BEV features $\bevfeaturemap$ that $\decoder$ produces. Hence, it is possible to apply this pretraining approach to any BEV model by plugging in the pretraining head, which we describe next, at the end of the BEV network, before the downstream task-specific head. 
We note that the auxiliary pretraining head is removed after the end of pretraining.

\subsection{Occupancy reconstruction} \label{sec:occupancy_reconstruction}

Building on the insights from previous studies that have highlighted the effectiveness of reconstruction as a valuable prior for diverse modalities, such as images~\cite{he2022masked, bao2022beit} and Lidar point clouds~\cite{also}, we employ a simple occupancy reconstruction pretraining task for BEV networks. 
The goal is to lead the BEV network to learn BEV features that encode information about the 3D geometry of the scene.

\begin{figure}
\makeatletter
\makeatother
\centering
\includegraphics[width=0.6\linewidth]{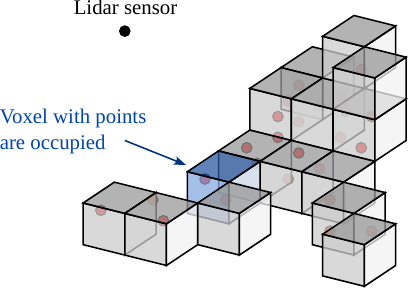}
\vspace{-9pt}
\caption{\textbf{Occupancy grid.}
A voxel is considered occupied if there is at least one point inside it.
}

\label{fig:occ_voxels}
\vspace{-15pt}
\end{figure}

Let $\voxgrid$ represent a voxel grid with shape $\bevzed \times \bevheight \times \bevwidth$, where $\bevzed$ is the height dimension. To create the occupancy targets $\occ \in \{0, 1\}^{\bevzed \times \bevheight \times \bevwidth}$, we adhere to the common convention~\cite{huang2023tri, tong2023scene} 
wherein a voxel $v \in \voxgrid$ is considered occupied if at least one Lidar point falls inside (see \cref{fig:occ_voxels}).

To estimate the occupancy grid $\hat{\occ}$, 
we employ an ``unsplatting'' decoder network $\lifting$ that takes the 2D BEV feature maps $\bevfeaturemap \in \realnum^{\numchannels \times \bevheight \times \bevwidth}$ from $\decoder$ as input and generates a 3D feature volume $\bevfeaturvolume \in \mathbb{R}^{\numchannels \times \bevzed \times \bevheight \times \bevwidth}$.

The unsplatting decoder $\lifting$ starts with two 2D convolutional layers. The first layer, with 3$\times$3 kernels and $\numchannels$ output channels, is followed by Instance Normalization~\cite{ulyanov2016instance} and ReLU units. The second layer has 1$\times$1 kernels with $\numchannels \bevzed$ output channels. These layers produce 2D BEV feature maps of shape $(\numchannels \bevzed) \times \bevheight \times \bevwidth$, which are reshaped into a 3D feature volume of shape $\numchannels \times \bevzed \times \bevheight \times \bevwidth$. This reshaping is done by dividing the $(\numchannels \bevzed)$-dimensional feature channels into $\bevzed$ groups, each with $\numchannels$ channels.

Next, the decoder $\lifting$ processes these 3D features with two 3D convolutional layers. The first layer, with 1$\times$1$\times$1 kernels and $2 \numchannels$ output channels, is followed by a Softplus non-linearity. The second layer has 1$\times$1$\times$1 kernels with $\numchannels$ output channels, producing the final 3D feature volume $\bevfeaturvolume \in \mathbb{R}^{\numchannels \times \bevzed \times \bevheight \times \bevwidth}$.

Finally, to generate the occupancy prediction $\hat{\occ}$, a single 3D convolution layer with a 1$\times$1$\times$1 kernel is applied on $\bevfeaturvolume$, followed by a sigmoid activation.

The loss function to minimize is a binary cross-entropy loss on the voxel occupation:
\begin{equation} \label{eq:occ_rec_loss}
\lossrec = \frac{1}{|\voxgrid|} \sum_{v \in \voxgrid} \text{BCE}(\hat{\occ}_v, \occ_v).
\end{equation}

\subsection{Occupancy-guided feature distillation} \label{sec:feature_distillation}

We introduce a self-supervised objective that complements occupancy reconstruction by guiding our BEV network to encode high-level semantic information. 
To achieve this, we leverage the availability of a  self-supervised pretrained image network, 
denoted as  $\teacherenc$, which takes an image as input and produces high-level 2D feature maps with $\numdistillchannels$ feature channels. 
Our \method~approach involves 
a feature distillation objective, where we fill the occupied voxels in $\voxgrid$ with features extracted from $\teacherenc$ and then train the BEV network to predict these voxel features.

Let $\occvox$ represent the set of occupied voxels, defined as $\occvox = \{\forall v \in \voxgrid \,|\, \occ_v = 1\}$. 
Our feature distillation objective operates specifically on these occupied voxels. 
To create the target feature $\teacherfeat_v \in \realnum^{\numdistillchannels}$  for each occupied voxel $v \in \occvox$, we project the voxel's center 3D coordinates onto the image feature maps extracted by the target image encoder $\teacherenc$ from the surround-view input images $\{I_c\}_{c=1}^{C}$.
Given these projections of 3D points into 2D images, we obtain an $N_y$-dimensional feature vector by bilinearly sampling a feature map of each image $I_c$ with a valid projection (i.e., if a point is projected inside an image $I_c$).
The target feature $\teacherfeat_v$ is then computed 
across images with valid projections as the average of the bilinearly-sampled feature vectors.
To predict the target features of the occupied voxels, we use the 3D feature volume $\bevfeaturvolume \in \mathbb{R}^{\numchannels \times \bevzed \times \bevheight \times \bevwidth}$ produced by the $\lifting$ decoder. 
A single 3D convolution layer with 1$\times$1$\times$1 kernels and $\numdistillchannels$ output channels is applied on $\bevfeaturvolume$, resulting in the 3D feature volume $\hat{\teacherfeat} \in \mathbb{R}^{\numdistillchannels \times \bevzed \times \bevheight \times \bevwidth}$. 
Then, for each occupied voxel $v \in \occvox$, we extract its corresponding feature $\hat{\teacherfeat}_v \in \realnum^{\numdistillchannels}$ from $\hat{\teacherfeat}$.

The feature distillation loss that we aim to minimize is the average negative cosine similarity between the predicted and target features for each occupied voxel $v \in \occvox$:
\begin{equation} \label{eq:distill_occ_loss}
\lossdistill = - \frac{1}{|\occvox|} \sum_{v \in \occvox} \text{cos}(\hat{\teacherfeat}_v, \teacherfeat_v).
\end{equation}

We note that there exists a small number of occupied voxels that lack valid projections into any of the images $\{I_c\}_{c=1}^{C}$.
Although not explicitly shown in \cref{eq:distill_occ_loss} for the sake of notation simplicity, these voxels are, in fact, excluded from the computation of the feature distillation loss.
 \section{Experiments}
\label{sec:experiments}

\subsection{Experimental setup}

\parag{Datasets.}
To evaluate our approach we use the nuScenes~\cite{nuscenes} dataset for both conducting the pretraining of the camera-only BEV networks and for finetuning them on the downstream tasks of BEV-based semantic segmentation.
The dataset is composed of 1,000 sequences 
recorded in Boston and Singapore.
The data is divided in training (700 sequences), validation (100 sequences) and test (200 sequences) splits.
Each frame contains a point cloud acquired with 32-layer Lidar and 6 images covering the surroundings of the ego-vehicle .

\parag{BEV segmentation tasks.}
We consider two different tasks related to BEV segmentation.
First, as in \cite{simplebev}, we evaluate our method on vehicle segmentation.
We build target BEV segmentation ground truth by projecting the boxes of vehicles on the BEV plane.
We use the same setting as \cite{simplebev}, i.e., a range of 50 meters around the ego-vehicle, and BEV ground truths of size 200x200 pixels.
Second, we focus on the layout and evaluate on map segmentation.
Here, we evaluate the ability of our finetuned BEV network to segment bakcground classes: ``road", ``sidewalk", ``crosswalk", ``parking area" and ``road dividers".

\newcommand{\nm}[1]{\textnormal{#1}}
\begin{table*}[ht!]
    \small
    \centering
    {
    \begin{tabular}{l|l|l|lll|lll}
        \toprule
         Architecture & Image & BEV & \multicolumn{3}{c|}{Vehicles} & \multicolumn{3}{c}{Map}\\
         & Backb. & Pretraining 
         & \multicolumn{1}{c}{1\%} 
         & \multicolumn{1}{c}{10\%} 
         & \multicolumn{1}{c|}{100\%} 
         & \multicolumn{1}{c}{1\%} 
         & \multicolumn{1}{c}{10\%} 
         & \multicolumn{1}{c}{100\%} \\
         \midrule
        && None 
        & 13.7
        & 26.0 
        & 37.4
        & 20.1
        & 28.1
        & 45.3 \\
         && \alsobaseline 
         & 18.5 \better{\Plus4.8}
         & 26.6 \worse{\Minus0.6}
         & 32.1 \worse{\Minus5.3}
         & 23.2 \better{\Plus3.1}
         & 29.0 \better{\Plus0.9}
         & 41.1 \worse{\Minus4.2}\\    
         \rowcolor{baselinecolor} \cellcolor{white} 
         &\cellcolor{white}
         &\method{} 
         & 24.3 \better{\Plus10.6} 
         & 30.9 \better{\Plus4.9}
         & 37.7 \better{\Plus0.3}
         & 25.4 \better{\Plus5.3}
         & 33.7 \better{\Plus5.6}
         & 47.3 \better{\Plus2.0}\\
        \rowcolor{baselinecolor} \cellcolor{white} 
        & \cellcolor{white} \multirow{-4}{*}{\effbzero}
        & \method$^\dagger$ 
        & 24.5 \better{\Plus10.8}
        & 32.0 \better{\Plus6.0}
        & 38.1 \better{\Plus0.7}
        & 26.2 \better{\Plus6.1}
        & 34.3 \better{\Plus6.2}
        & 47.6 \better{\Plus2.3} \\
         \cmidrule{2-9}
        && None 
        & 13.9
        & 28.6
        & 42.2
        & 20.1
        & 31.2
        & 44.2\\
        &&\alsobaseline 
         & 18.4 \better{\Plus4.5}
         & 25.6 \worse{\Minus3.0}
         & 37.2 \worse{\Minus5.0}
         & 22.3 \better{\Plus2.2}
         & 28.7 \worse{\Minus2.5}
         & 41.2 \worse{\Minus3.0}\\    
          \rowcolor{baselinecolor} \cellcolor{white} \multirow{-8}{*}{SimpleBEV}
         & \cellcolor{white}\multirow{-4}{*}
         &\method{}
         & 23.5 \better{\Plus9.6}
         & 31.4 \better{\Plus2.8} 
         & 41.0 \worse{\Minus1.2}
         & 25.2 \better{\Plus5.1}
         & 33.6 \better{\Plus2.4}
         & 50.0 \better{\Plus 5.8}  \\
          \rowcolor{baselinecolor} \cellcolor{white} 
         & \cellcolor{white}\multirow{-4}{*}{\resnetfifty}
         &\method{}$^\dagger$
         & 24.8 \better{\Plus10.9}
         & 31.3 \better{\Plus2.7}
         & 41.7 \worse{\Minus0.5}
         & 25.9 \better{\Plus5.8}
         & 34.4 \better{\Plus3.2}
         & 51.3 \better{\Plus7.1}\\
    \midrule
        &&None           
            & 11.3
            & 22.8  
            & 37.2
            & 20.4 
            & 29.4 
            & 49.6 
            \\
         \rowcolor{baselinecolor} \cellcolor{white} \multirow{-2}{*}{BEVFormer}
         & \cellcolor{white} \multirow{-2}{*}{\effbzero}  
         & \method{} 
         & 24.9 \better{\Plus13.6}
         & 30.1 \better{\Plus7.3}
         & 38.8 \better{\Plus1.6}
         & 26.6 \better{\Plus6.2}
         & 36.2 \better{\Plus6.8}
         & 50.5 \better{\Plus0.9} \\

    \bottomrule
    \end{tabular}
    }
    \vspace{-5pt}
    \caption{\textbf{Segmentation IoU results for Vehicle and Map classes}. Comparing our \method~against no BEV pretraining (None) and 
    the 
    pretraining baseline \alsobaseline. Results with 224$\times$400 image resolution.
    $^\dagger$: pretrained for 100 epochs; 
    all other models for 50 epochs.}
    \label{tab:comparison_baselines}
    \vspace{-14pt}
\end{table*}

\parag{Architectures.}
\noindent{\emph{BEV networks}.} 
We experiment with two BEV architectures: SimpleBEV~\cite{simplebev} and BEVFormer~\cite{li2022bevformer} (implementation from~\cite{simplebev}).

\noindent{\emph{Image encoders}.} 
As image backbones in the BEV segmentation networks, we employ either EfficientNet-B0~\cite{effnet} (\effbzero) or ResNet-50~\cite{resnet} (\resnetfifty). Following the common practice in camera-only BEV segmentation \cite{simplebev}, these image backbones have undergone pretraining on ImageNet, either in a supervised way (for \effbzero) or self-supervised using MoCov2~\cite{chen2020improved} (for \resnetfifty). 

\noindent{\emph{Teacher model.}}
The self-supervised image foundation model $\teacherenc$, used as the teacher for the occupancy-guided feature distillation, is the ViT-S/14 variant of DINOv2~\cite{dinov2}.

\parag{\method~pretraining.}
We pretrain BEV segmentation networks with batch size 16 on 4$\times$V100 GPUs using the Adam optimizer with weight decay $1e{-}7$ and a constant learning rate of $1e{-}3$ during training.
By default we pretrain for 50 epochs. 
We also pretrain some models for 100 epochs, which we denote with \method$^\dagger$ in the tables.
We use augmentations from \cite{simplebev}. Except otherwise stated, we use 224$\times$400 image resolution for the input camera frames.

\parag{Finetuning.}
We experiment with finetuning at 1\%, 10\% and 100\%. We  use the the AdamW optimizer and One Cycle scheduler with $1e{-}7$ as weight decay. The complete hyperparameters (number of epochs / starting learning rate / batch size / batch gradient accumulation) are the following, 1\%: 100/$1e{-}4$/6/1; 10\%: 50/$1e{-}4$/6/1 and 100\%: 30/$3e{-}4$/8/5. The training is conducted with 2$\times$V100 GPUs.
If the BEV segmentation network undergoing finetuning has not been pretrained with \method, we initialize its image backbone with 
weights pretrained on ImageNet, following the same initialization procedure used before the \method~pretraining.
This is common practice in camera-only BEV segmentation \cite{simplebev}.

\subsection{Pretraining baselines}
To validate the effectiveness of our \method~approach, which integrates feature distillation with 3D geometry prediction, we implement a pretraining baseline focusing solely on 3D geometry reconstruction.
To implement this baseline, called \alsobaseline, 
we modify the Lidar pretraining method ALSO \cite{also} for the purpose of pretraining camera-only BEV networks. ALSO is a top-performing reconstruction-based approach for self-supervised pre-training Lidar networks. Its pretext task is to learn an implicit function partitioning the space between empty and occupied (inside objects) spaces, where the supervision signal is directly generated from the input Lidar data, using the sensor line of sights. This entails a geometry reconstruction task, where each pair of 3D point/output feature must reconstruct a local neighborhood. To implement our \alsobaseline~baseline, we begin with the ALSO variant tailored for pretraining Lidar detection networks, such as the SECOND network~\cite{Yan2018SECONDSE}, where the features from the input 3D point cloud are projected on a top-view plane. In this setup, we replace the 3D backbone with the image BEV network intended for pretraining. As the supervision signal in \alsobaseline~still comes from the Lidar points (as in ALSO), we use the same hyper-parameters as ALSO, i.e., decimation grid of 10cm, maximal distance-to-Lidar-point for query of 10cm and a reconstruction radius of 1m. We compare against \alsobaseline~in \cref{sec:results}.

In addition to comparing \method~with the above baseline, in \cref{sec:ablation_study} we include comparisons against ablations that use only the occupancy reconstruction objective ($\lossrec$), which is another 3D geometry-based approach, or only the occupancy-guided feature distillation objective ($\lossdistill$). 

\subsection{Results} \label{sec:results}

\parag{Comparison with baselines.} \label{sec:comparison_with_baselines}
In the SimpleBEV results presented in \cref{tab:comparison_baselines}, our \method~self-supervised BEV pretraining approach is compared against the proposed baseline \alsobaseline, 
along with the case of not conducting BEV pretraining. We provide results for both Vehicle and Map segmentation, utilizing 1\%, 10\%, and 100\% of annotated training data. For the SimpleBEV network we use either the \effbzero~or the \resnetfifty~image backbones and 224$\times$400 image resolution for the input camera frames.
 
We observe that, compared to the absence of BEV pretraining, our \method~enhances segmentation results in almost all settings.
The only exception is that of Vehicle segmentation with 100\% annotations using SimpleBEV with \resnetfifty. 
The improvement is particularly prominent with only 1\% or 10\% of annotated data, showcasing the effectiveness of our approach in low-shot settings. When comparing with the examined baseline (\alsobaseline), our \method~outperforms them in almost all cases. An interesting observation is that \alsobaseline~improves segmentation performance only in the 1\%-annotation settings, while yielding worse results for the 10\%- and 100\%-annotation settings.

\begin{figure}[t!]
    \centering
    \includegraphics[width=0.8\linewidth]{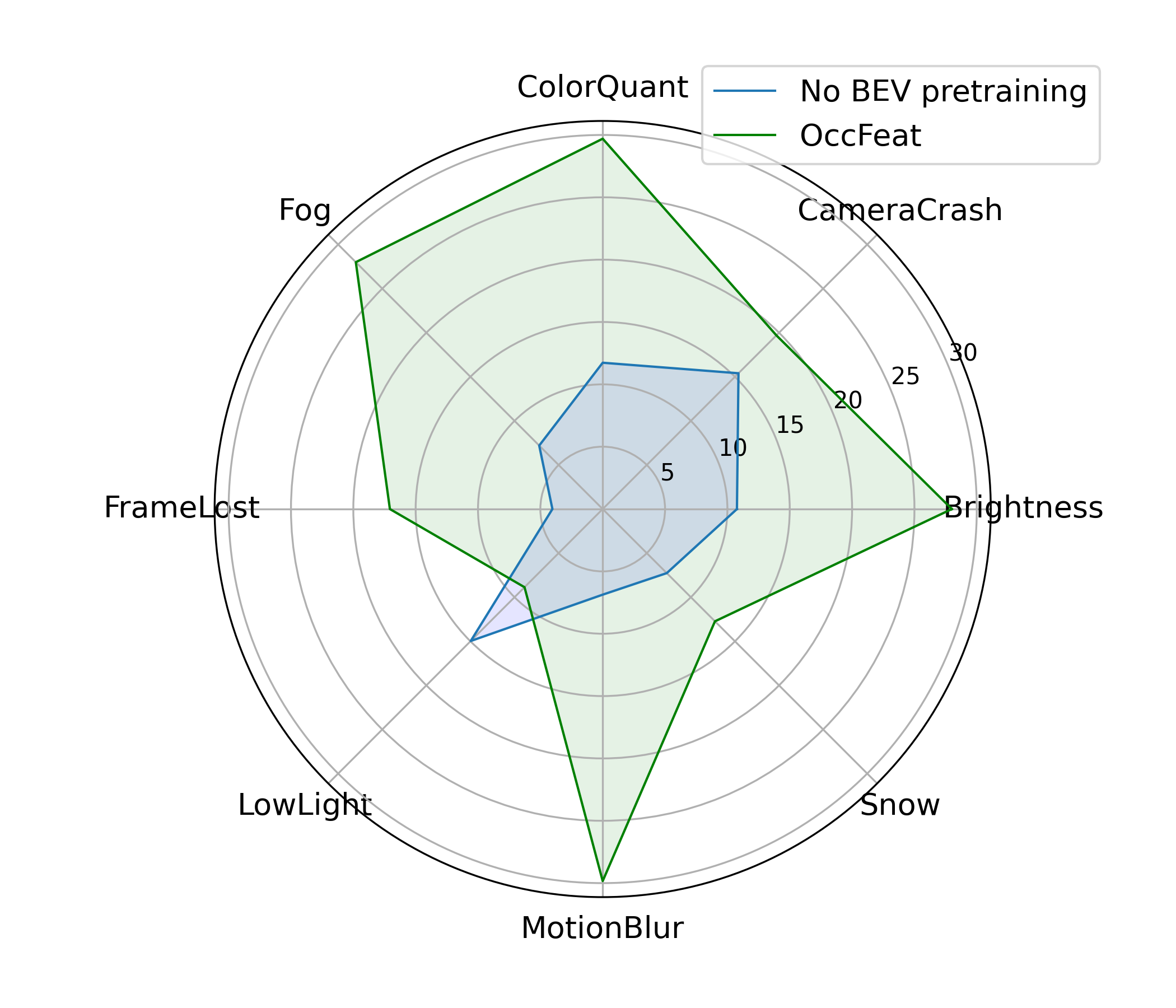}
    \vspace{-17pt}
    \caption{\textbf{Study on robustness}. Segmentation results on nuScenes-C dataset for Vehicle classes using BEVFormer network with EN-B0 image backbone on 100\% annotated data. Comparison of our OccFeat against no BEV pretraining.}
    \label{fig:nuscenesc}
    \vspace{-5pt}
\end{figure}

\parag{Scaling with longer BEV pretraining.}
In \cref{tab:comparison_baselines}, apart from presenting results for 50 pretraining epochs, we also include results for our \method~approach and SimpleBEV (with \effbzero~or \resnetfifty) 
with an extended pretraining duration of 100 epochs (\method$^\dagger$ model). We observe consistent improvements in segmentation results with this longer pretraining duration. This observation underscores the scalability of our approach, demonstrating its ability to gain further benefits from longer pretraining periods.

\parag{Adaptable to various BEV network architectures.}
As mentioned in \cref{sec:method}, our BEV pretraining method can be used with any BEV network architecture. 
In \cref{tab:comparison_baselines}, apart from results with SimpleBEV networks, we also present results using the BEVFormer \cite{li2022bevformer} segmentation network for the 1\%, 10\%, and 100\% annotation settings.
We compare the performance of our \method~method against the scenario where BEV pretraining is not conducted. We see that even when employing the BEVFormer network architecture, our \method~pretraining approach enhances segmentation performance across all evaluation settings.

\begin{table}[t]
\small
\centering
\setlength{\tabcolsep}{4pt}
\resizebox{0.9\linewidth}{!}{
\begin{tabular}{l|c|c|lll}
\toprule
\multirow{2}{*}{BEV Pretraining} & \multicolumn{2}{c|}{Image Resolution} & \multicolumn{3}{c}{Vehicles}\\ 
& Pretraining & Finetuning   & \multicolumn{1}{c}{1\%} & \multicolumn{1}{c}{10\%} & \multicolumn{1}{c}{100\%}\\
\midrule
None & N/A & \multirow{3}{*}{224$\times$400} & 13.7 & 26.0 & 37.4\\
\method & 224$\times$400 &  & 24.3 & 30.9 & 37.7\\
\method$^\dagger$ & 224$\times$400 &  & 24.5 & 32.0 & 38.1\\
\midrule
None & N/A & \multirow{3}{*}{448$\times$800} & 12.9 & 28.4& 41.6\\
\method & 448$\times$800 &  & 26.5 &34.5& 41.5\\
\method$^\dagger$ & 224$\times$400 &  & 26.1& 33.3 &41.5\\
\bottomrule
\end{tabular}
}
\caption{\textbf{Impact of image resolution}.
SimpleBEV segmentation IoU results for the Vehicle class.
Results with the \effbzero~image backbone.
$^\dagger$: pretrained for 100 epochs.
}
\label{tab:img_resolution}
\vspace{-12pt}
\end{table}
\begin{table}[t]

    \small
    \centering
    \setlength{\tabcolsep}{2pt}
      \resizebox{0.9\linewidth}{!}{
    \begin{tabular}{c| c | c c | c | c }
        \toprule
         \multirow{2}{*}{BEV Network} & BEV & \multicolumn{2}{c|}{\method~losses} & Vehicles   & \,\,\,Maps\,\,\, \\
         & Pretraining & \,\,\,$\lossrec$\,\,\, & $\lossdistill$ & 1\% &  1\% \\
         \midrule
         \multirow{4}{*}{SimpleBEV} & \xmark & & & 13.7 & 20.1 \\
         & \cmark & \cmark & & 22.8 & 25.2 \\
         & \cmark & & \cmark & 17.4 & 23.4\\
         & \baseline{\cmark} & \baseline{\cmark} & \baseline{\cmark} & \baseline{24.3} & \baseline{25.4}\\
        \midrule
         \multirow{3}{*}{BEVFormer} & \xmark & & & 11.3 & 20.4 \\
         & \cmark & \cmark & & 21.9 & 25.1 \\
         & \baseline{\cmark} & \baseline{\cmark} & \baseline{\cmark} & \baseline{24.9} & \baseline{26.6}\\         

         \bottomrule
    \end{tabular}
    }
    \caption{\textbf{Ablation of \method's losses}. BEV segmentation results (IoU) for the Vehicle and Map classes using 1\% annotated data. Results with 224$\times$400 resolution and the \effbzero~image backbone.}
    \vspace{-5pt}
    \label{tab:ablation_losses}
\end{table}
\begin{table}[t]
    \small
    \setlength{\tabcolsep}{4pt}
    \centering
    \resizebox{0.9\linewidth}{!}
    {
    \begin{tabular}{l|cccccc} 
    \toprule
    Loss weight $\lambda$   & 0    & 0.0001 & 0.001 & \baseline{0.01} & 0.1 & 1.0\\
    \midrule
    Vehicle (1\%) & 22.8&22.9   & 22.7  & \baseline{\textbf{24.3}} & 23.9& 19.4\\
    \bottomrule
    \end{tabular}
    }
    \caption{\textbf{Impact of loss weight $\lambda$} ($\losstot = \lossrec + \lambda \cdot \lossdistill$). 
    SimpleBEV vehicle segmentation results using 1\% annotated data, 224$\times$400 image resolution and the \effbzero~image backbone.}
    \vspace{-12pt}
\label{tab:loss_weight}
\end{table}

\parag{Exploiting higher resolution images.}
In \cref{tab:img_resolution}, we present segmentation results using either 224$\times$400 or 448$\times$800 resolutions for the camera frames fed into the SimpleBEV network. 
Our \method~pretraining approach consistently enhances segmentation results across all cases. 
What is noteworthy is that pretraining with the lower 224$\times$400 resolution and then fine-tuning with the higher 448$\times$800 resolution also improves the results, almost as much as pretraining directly with the higher resolution. 

\parag{Robustness study.} We study the robustness of \method{} by evaluating it on the nuScenes-C benchmark~\cite{xie2023robobev}.
This benchmark consists of eight distinct data corruptions,
each with three intensity levels, applied to the validation set of nuScenes.
In \cref{fig:nuscenesc} we 
present vehicle segmentation results on nuScenes-C using BEVFormer with \effbzero~backbone finetuned on 100\% annotation data.
For each corruption type we report the average across three severity levels. The comparison of our \method{} against no BEV pretraining illustrates that the \method{} pretraining improves the robustness of the final BEV model.

\subsection{Ablation study} \label{sec:ablation_study}

\begin{figure*}[t]
    \centering

    \includegraphics[width=0.99\linewidth]{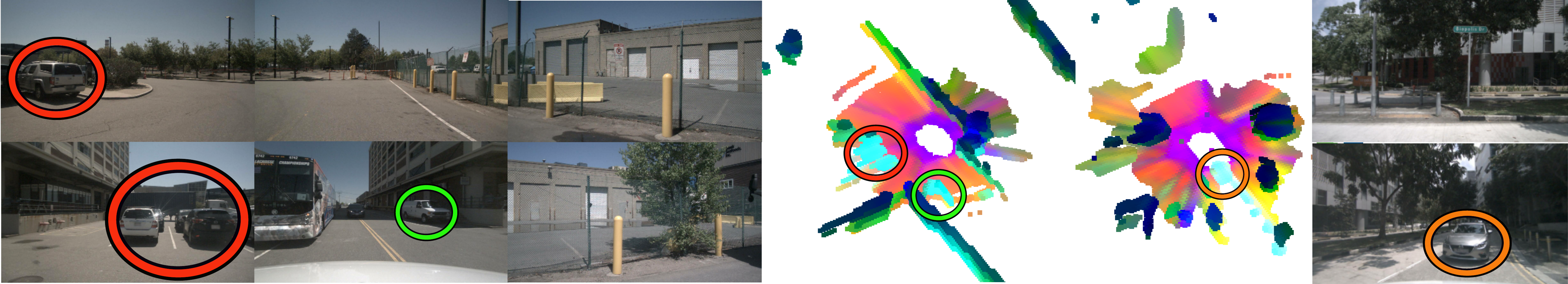}

\vspace{-5pt}
    \caption{\textbf{Visualisation of predicted 3D features,} using a 3-dimensional PCA mapped on RGB channels. The features contain semantic information, e.g., cars in cyan color. Using the same PCA mapping on a different scene (right), we show that semantic features are consistent across scenes. Objects within colored circles in the feature space correspond to those in similarly colored circles in the image.} 
    \label{fig:3d_features}

    \vspace{-10pt}
\end{figure*}

\parag{Ablation of \method's losses.} 
In \cref{tab:ablation_losses} we conduct an ablation study on the two pretraining objectives, namely $\lossrec$ and $\lossdistill$, of our \method~approach. The evaluation focuses on Vehicle and Map segmentation results with 1\% annotated training data, using both the SimpleBEV and BEVFormer networks. For SimpleBEV, both $\lossrec$ (a 3D geometry prediction objective) and $\lossdistill$ (an occupancy-guided feature distillation objective) showcase improvements compared to the scenario without BEV pretraining. The combination of both pretraining objectives, forming our \method~approach, yields the most favorable segmentation results. This underscores the efficacy of our \method~pretraining method, distinguishing it from prior self-supervised BEV pretraining works that rely solely on 3D geometry prediction. 
The advantage of enhancing 3D geometry prediction ($\lossrec$) with feature distillation ($\lossdistill$) is further affirmed by the ablation results obtained with BEVFormer.

\parag{Impact of loss weight $\lambda$.} 
In \cref{tab:loss_weight}, we investigate the influence of the loss weight $\lambda$, responsible for balancing the two loss terms of \method~(refer to \cref{eq:total_loss}). The most favorable segmentation results are achieved for $\lambda$ values ranging between 0.001 and 0.01, with 0.01 being the optimal choice.

\begin{figure}[t!]
  \centering
  \begin{subfigure}[b]{0.45\linewidth}
    \includegraphics[width=\linewidth, trim={21cm 0 0 0}, clip]{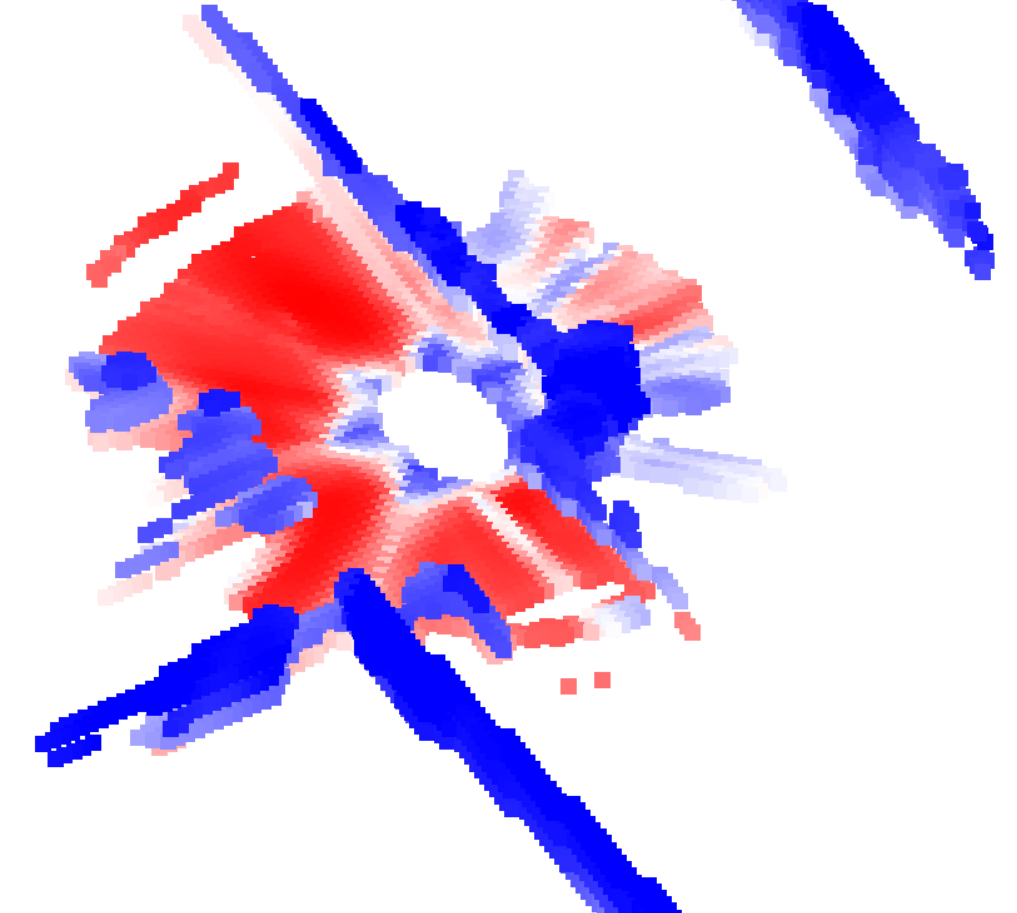}
    \caption{Road}
    \label{fig:car}
  \end{subfigure}
  \hfill
  \begin{subfigure}[b]{0.45\linewidth}
    \includegraphics[width=\linewidth, trim={0 0 21cm 0}, clip]{figures/correlationsimg.pdf}
    \caption{Car}
    \label{fig:road}
  \end{subfigure}
  \vspace{-8pt}
    \caption{
    \textbf{Correlation maps} of the student's predicted 3D features and features selected on the road (a) and on a car (b).}
  \label{fig:correlations}
  \vspace{-10pt}
\end{figure}

\subsection{Qualitative results}

In~\cref{fig:3d_features} and~\cref{fig:correlations}, we assess the semantic quality of the reconstructed features, using a colored mapping from PCA (\cref{fig:3d_features}) and correlation maps (\cref{fig:correlations}). We can observe that semantic information from DINOv2 teacher has been preserved and semantic classes such as cars are easily separable.
Additionally, the representations are consistent accross scenes, e.g., on~\cref{fig:3d_features} (right), we apply the PCA mapping computed on the left scene to a new scene. As an example, the cyan points on the right correspond to a car.

\section{Conclusion}
\label{sec:conclusion}

We introduced \method, a self-supervised pretraining method for camera-only BEV segmentation networks. Our approach combines two pretraining objectives: a 3D occupancy prediction task using raw Lidar data and an occupancy-guided feature distillation task based on the self-supervised pre-trained image foundation model DINOv2. The former enhances the learning of 3D geometry-aware BEV features, while the latter focuses on semantic-aware BEV features. Our empirical results demonstrate that both pretraining objectives enhance segmentation performance compared to not conducting BEV pre-training. Combining both objectives yields the most favorable results, emphasizing the effectiveness of our \method~approach. This sets our method apart from prior self-supervised BEV pre-training methods that solely rely on 3D geometry prediction.

\vspace{-6pt}
\paragraph{Limitations and Perspectives.} While \method{} has proven highly effective for low-data scenarios (i.e., 1\% and 10\% finetuning), it yields slight or no improvements in the 100\% finetuning setting. This could be because we self-supervisedly pretrained and then supervisedly finetuned on the same data, leaving all information available at finetuning. 
Additionally, the nuScenes dataset used for pretraining is relatively small. Perhaps \emph{using a larger pretraining dataset} could further enhance performance in the 100\% finetuning setting.

One aspect not sufficiently explored in our work is \emph{``scaling the pretraining"}. As demonstrated in ScaLR~\cite{puy24scalr}, when pretraining Lidar networks via self-supervised distillation of image-based models, scaling the teacher and student can significantly boost performance. For example, using a larger teacher model, such as going from ViT-S to ViT-L or ViT-G variants of DINOv2, could yield superior features for \method's distillation loss. Similarly, scaling the student components —specifically, the image encoder and BEV decoder—might enhance \method's distillation process.

Furthermore, another avenue for improvement could involve \emph{incorporating time} into our self-supervised pretraining task. For instance, accumulating Lidar points over several frames to generate denser occupancy maps could enhance the effectiveness of \method's pretraining.

\footnotesize{\paragraph{\footnotesize{Acknowledgements}}
This work was performed using HPC resources from GENCI–IDRIS (Grants 2022-AD011012883R2, 2022-AD011012884R2 and 2024-AD011015037). 
It was supported in part by the French Agence Nationale de la Recherche (ANR) grant MultiTrans (ANR21-CE23-0032).
It also received the support of CTU Student Grant SGS21184OHK33T37 and by the Ministry of Education, Youth and Sports of the Czech Republic through the e-INFRA CZ (ID:90254), and 
the support of EXA4MIND, a European Union’s Horizon Europe Research and Innovation programme under grant agreement N°101092944. Views and opinions expressed are however those of the author(s) only and do not necessarily reflect those of the European Union or the European Commission. Neither the European Union nor the granting authority can be held responsible for them. The authors have no competing interests to declare that are relevant to the content of this article.}

\clearpage

{
    \small
    \bibliographystyle{ieeenat_fullname}
    \bibliography{main}
}


\end{document}